\documentclass[sigconf]{acmart}

\usepackage{booktabs} 
\usepackage{graphicx} 
\usepackage{subfigure} 
\usepackage{natbib}
\usepackage{algorithm}
\usepackage{algorithmic}
\usepackage[colorinlistoftodos]{todonotes}
\usepackage{rotating}
\usepackage{hyperref}
\usepackage{siunitx}
\usepackage{amsmath}

\usepackage{xcolor,colortbl} 

\setcopyright{none}

\acmISBN{}
\acmPrice{}
\acmDOI{}
\acmYear{}

\begin{document}
\title{
Sequential Attacks on Agents for Long-Term Adversarial Goals
\vspace{1em}
}

\author{Edgar Tretschk$^{1,2}$\quad\quad\quad\quad\quad\quad Seong Joon Oh$^{1}$\quad\quad\quad\quad\quad\quad Mario Fritz$^{1}$}

\affiliation{\vspace{1em}$^{1}$Max-Planck Institute for Informatics, Saarland Informatics Campus, Germany}
\affiliation{$^{2}$Saarland University, Germany}
\affiliation{\{tretschk,joon,mfritz\}@mpi-inf.mpg.de}

\renewcommand{\shortauthors}{E. Tretschk et al.}

\begin{abstract}
Reinforcement learning (RL) has advanced greatly in the past few years with the employment of effective deep neural networks (DNNs) on the policy networks~\cite{krizhevsky2012imagenet,mnih2015human}. With the great effectiveness came serious vulnerability issues with DNNs that small adversarial perturbations on the input can change the output of the network~\cite{szegedy2014iclr}. Several works have pointed out that learned agents with a DNN policy network can be manipulated against achieving the original task through a sequence of small perturbations on the input states~\cite{HuangPGDA17}. In this paper, we demonstrate furthermore that it is also possible to impose an arbitrary adversarial reward on the victim policy network through a sequence of attacks. Our method involves the latest adversarial attack technique, Adversarial Transformer Network (ATN)~\cite{baluja2017adversarial}, that learns to generate the attack and is easy to integrate into the policy network. As a result of our attack, the victim agent is misguided to optimise for the adversarial reward over time. Our results expose serious security threats for RL applications in safety-critical systems including drones~\cite{gandhi2017learning}, medical analysis~\cite{nemati2016optimal}, and self-driving cars~\cite{sallab2017deep}. 
\end{abstract}

\maketitle

\section{Introduction}
\label{sec:1intro} 

Recent years have seen great advances in reinforcement learning (RL). Owing to successful applications of deep neural networks on policy networks~\cite{mnih2015human}, RL has surpassed human-level performance in Atari games~\cite{mnih2015human} and in the game of Go~\cite{silver2016mastering}. It is seeking its way into security-critical applications like self-driving cars~\cite{sallab2017deep}.

While being effective, deep neural networks are known to be vulnerable to adversarial examples, small perturbations on the input that make the network confidently predict wrong outputs~\cite{szegedy2014iclr}. Huang et al.~\cite{HuangPGDA17} have shown that deep policy networks are no exceptions. A sequence of small perturbations on the environment (Atari game pixels) result in the agent performing significantly worse on the given task. Many follow-up works have investigated further vulnerabilities of deep policy networks~\cite{LinHLSLS17,kos2017delving} and proposed strategies to make them more robust~\cite{BehzadanM17,lin2017detecting,mandlekar2017adversarially,pinto2017robust,pattanaik2017robust}.

In this work, we show that a sequence of small attacks on the deep policy network can not only make the agent underperform on the original task, but also manoeuvre it to \emph{pursue an adversarial goal}. For example, given a self-driving vehicle trained to transport goods from a seaport to a sorting centre, an adversary applies a sequence of perturbations on the vehicle's sensor to deliver the goods to the adversary's property, without altering the vehicle's policy network. Such an attack would be more appealing to the adversary than simply making the agent fail. The attack we consider is therefore realistic and relevant.

We build our threat model as a perturbation network, which, together with the victim's policy network, becomes a policy network that pursues the adversarial goal.
Specifically, assume that agent~$a$ follows the policy network $f$ to maximise the original reward $r^O$ in the long term. Our threat model is represented as a feed-forward adversarial transformer network (ATN)~\cite{baluja2017adversarial} $g:X\rightarrow X$. $g$ produces small perturbations over input sequences such that the agent's policy network over the perturbed inputs $f(x+g(x))$ pursues the adversarial reward $r^A\neq r^O$. For making the perturbations small, we project the adversarial perturbation $g(x)$ onto the $\ell_2$ ball of radius $p$.

We train and evaluate our threat model over agents trained for the Pong Atari game. Our adversaries successfully make the agents pursue the new, adversarial goal (hitting the centre $1/5$ region in the enemy's score line) through a sequence of quasi-perceptible perturbations over the input pixels. 

This paper contributes the following: (1) a threat model that generates a sequence of perturbations that manoeuvre a policy network to pursue an adversarial reward at test time and (2) empirical evidence that the suggested threat model successfully achieves the adversarial goal. Our work exposes crucial yet previously unseen security risks of real-life deployment of RL based agents.

\section{Related Work}
\label{sec:2related} 

We describe prior work in three relevant areas: (1) reinforcement learning, (2) machine learning vulnerability, and (3) vulnerability of deep policy networks. Our work will be discussed in the context of those prior literature.

\subsection{Reinforcement Learning}
\label{subsec:2RL}

Reinforcement learning (RL) enables agents to learn by interacting with the environment to achieve long-term rewards. The advantage of not requiring per-action supervision has attracted much research in the field in application areas that involve long-term and complex action-reward structures: boardgames~\cite{tesauro1995td}, inverse pendulum~\cite{bertsekas1995dynamic}, and robotics~\cite{peters2003reinforcement}.

Development of highly performant deep neural networks (DNN)~\cite{krizhevsky2012imagenet} has trickled down to the effective deep policy networks (Deep Q-Networks, DQN) for RL. Mnih et al.~\cite{mnih2015human} have first applied the DQNs to learn to play 43 diverse Atari games, and have demonstrated super-human performances. The seminal work by Silver et al.~\cite{silver2016mastering} has showcased the ability of a DQN-based agent for playing the game of Go to overwhelm human experts.

Deep reinforcement learning is an active area of research. Many improvements to the algorithms have been proposed in the last years. 
Double Q-learning \cite{van2016deep} and dueling networks \cite{wang2016dueling} were significant steps forward for the usability of DQNs. Among the most widely used improvements is also prioritized replay \cite{schaul2015prioritized}, which we employ in our method. 
As a public contribution, OpenAI has open-sourced the Atari game environments (OpenAI Gym,~\cite{openaigym}). Our experiments on the game of Pong are built on the baseline implementation of DQN by OpenAI~\cite{openAIbaselines}. We also use the pre-trained Pong agents from \cite{openAIbaselines}.

\subsection{Attacking Machine Learning Models}
\label{subsec:2Attacks}

While fragility of learned models has been studied for a long time~\cite{Lanckriet2003minimax,lowd2005adversarial,kolcz2009feature}, it has received more attention in the recent years after deep neural networks were found to be vulnerable to human-imperceptible adversarial perturbations~\cite{szegedy2014iclr}.

\subsubsection{Victim Models}

Most frequently used victim models for adversarial attack research are classification models: given an input, predict the corresponding class~\cite{goodfellow2015iclr,moosavi2016cvpr,moosavi2016universal,joon17iccv}. Other works have verified the vulnerability of models for generative~\cite{kos2017adversarial}, and detection and segmentation~\cite{xie2017adversarial} tasks.

Huang et al.~\cite{HuangPGDA17} first showed that deep neural networks are vulnerable for reinforcement learning tasks. Our work also studies the model vulnerability in the RL setup. We will compare our work against \cite{HuangPGDA17} and follow-up works in \S\ref{subsec:2attackRL}.

\subsubsection{Targeted Versus Non-Targeted Attacks}

Researchers have considered two types of adversarial attacks against models: ones inducing \emph{any} change in prediction (non-targeted) and ones inducing a \emph{specific} prediction (targeted)~\cite{goodfellow2015iclr,dong2017discovering}. This work considers an analogue of targeted attack in the reinforcement learning setup. Our attacks can not only make an agent fail, but also make it actively pursue an adversarial goal.

\subsubsection{Attack Algorithms}

Since the first discovery of the imperceptible adversarial examples~\cite{szegedy2014iclr}, researchers have developed more efficient, more efficient, and more resilient adversarial perturbation algorithms~\cite{goodfellow2015iclr,moosavi2016cvpr,moosavi2016universal,kurakin2016adversarial,joon17iccv}.
In particular, Baluja et al.~\cite{baluja2017adversarial} have proposed the Adversarial Transformation Networks (ATN).
Unlike prior works that generate perturbations by computing gradients $\nabla_x f(x)$ from the target network $f$, ATN is a learned function $g$ that transforms the input $x$ into an adversarial perturbation $g(x)$ such that the victim network $f$ is fooled when $x+g(x)$ is given.

\subsection{Attacking Agents}
\label{subsec:2attackRL}

Huang et al.~\cite{HuangPGDA17} have shown for the first time that deep policy networks are also susceptible to adversarial perturbations; small perturbations that would not interfere with human performance have significantly reduced the test time reward for the agents in various Atari game environments. They have further verified that the perturbations transfer across agents pursuing the same task. 
Independently, Kos et al.~\cite{kos2017delving} have also proposed adversarial attacks that reduce the test time rewards. Unlike \cite{HuangPGDA17} that attacks the agent on every frame, they have considered timing attacks where attacks are performed intermittently. Many follow-up works have expanded the research frontier in different directions. 
 
Researchers have considered injecting adversarial perturbations on the environments during training to learn policy networks that are more robust at test time.
Pinto et al.~\cite{pinto2017robust} and Pattanaik et al.~\cite{pattanaik2017robust} have suggested a minimax training objective for the agent, where an add-on adversary continually injects reward-minimising changes on the environment. As a result of this training, they have reported better generalisation and robustness against adversarial attacks at test time.

In contrast to this line of work where the adversary only strives to make the agent fail on the original task, a few studies have considered driving the victim towards a certain state or goal.
Lin et al.~\cite{LinHLSLS17} have proposed the \emph{enchanting attack} in which the adversary sequentially perturbs the input states (frames) $s_t$ for time steps $t=0, \cdots, H-1$ to guide the agent towards a predefined adversarial state $s_A$ at time $t=H$. While sharing similarities, our adversary imposes an adversarial \emph{reward} $r^A$ on the victim, instead of an adversarial state $s_A$; the former is more flexible and can encode the latter. 

Behzadan et al.~\cite{BehzadanM17} have proposed the \emph{Policy Induction Attacks}. In this attack, the adversary first trains a policy network (DQN) with an adversarial reward $r^A$. Using the trained policy, it crafts a sequence of \emph{targeted} adversarial perturbations that lead the victim's DQN to a sequence of actions leading to $r^A$. 
While related, their attacks are applied during training to make the agent learn the adversarial reward. Our attacks, on the other hand, are applied at test time and do not explicitly model a secondary DQN for planning actions; we train a feed-forward state perturbation module that is added on the input stream of the victim DQN. We adapt their method as a baseline to our setting.

\section{Background}
\label{sec:3background}

In this section, we provide background on the reinforcement learning (RL) setup and techniques, and adversarial attacks in general.

\subsection{Reinforcement Learning}
\label{subsec:3RL}

The RL agent is assumed to be interacting with the environment through a Markov Decision Process (MDP)~\cite{bellman1957markovian}, specified by the 5-tuple $(S,A,T,R,\gamma)$, where $S$ and $A$ are the state and action spaces, $T$ is the transition model, $R$ is the reward for the agent, and $0 <\gamma< 1$ is the discount factor. An MDP is a stochastic process for $t\geq 0$ that depends on the agent's action sequences $a_0,a_1,\cdots$. Specifically, given a state $s_t\in S$ and action $a_t\in A$ taken by the agent, the next state is determined stochastically by the transition model $s_{t+1}\overset{d}{\sim}T(s_t,a_t)$, and the reward is given by $R(s_t,a_t)$. The goal of RL is to find the optimal policy $\pi:S\rightarrow A$ that maximises the discounted reward
\begin{align}
\sum_{t\geq 0} \gamma^t R(s_t, \pi(s_t)).
\end{align}

One of the most promising approaches to this problem is the Q-learning paradigm, also referred to as backward induction~\cite{bellman1957markovian}. We define an auxiliary Q-function $Q:S\times A\rightarrow \mathbb{R}$ that returns the discounted future reward attained, given a state-action pair at time $t$, and following the optimal policy afterwards. Once we have access to the Q-function, we can obtain the optimal policy by computing $\pi(s)=\underset{a\in A}{\arg\!\max}\,Q(s,a)$.

In Q-learning, $Q$ is initialised randomly, and then approximated by sequential Bellmann updates~\cite{bellman1957markovian}:
\begin{align}
Q(s_t,a_t)\leftarrow 
R(s_t,a_t)+\gamma\cdot\underset{a}{\max}\,Q(T(s_t,a_t),a)
\end{align}
A Deep Q-Learning Network (DQN) models $Q$ as a deep neural network $f$ that takes the state $s_t$ as input and returns a vector of scores over the actions $a_t$ as output~\cite{mnih2015human}. The training objective is given by 
\begin{align}
\label{eq:DQN-objective}
\underset{\phi}{\min}\,\,\underset{s_t,a_t}{\mathbb{E}}\left[\left(
y-Q_\phi(s_t,a_t)
\right)^2\right]
\end{align}
where $y:=R(s_t,a_t)+\gamma\cdot\underset{s}{\mathbb{E}}\,[\underset{a}{\max}\,Q_{\phi^\prime}(T(s_t,a_t),a)]$
is the target Q~value computed separately via a target DQN parametrized by~$\phi^\prime$. Periodically, $\phi^\prime$ is set to $\phi$ and then kept fixed again. This improves training stability. During training, exploration of the state space yields obervation tuples $(s_t, a_t, T(s_t,a_t), R(s_t,a_t))$. These are stored in a replay buffer and later used to approximate the training objective. Prioritized replay \cite{schaul2015prioritized} implements this replay buffer as a priority queue, with priorities set to the temporal difference error~$\vert y-Q_\phi(s_t,a_t)\vert$.

Applying deeply learned policy networks has led to many breakthroughs in performances for RL. In this paper, we consider attacking DQNs by imposing an adversarial policy through sequential, small perturbations on the states $s_t$ at test time.

\subsection{Adversarial Attacks}
\label{subsec:3attack}

While deep neural networks have enjoyed super-human performances in various tasks, including reinforcement learning, they have been found to be susceptible to small (in the range between imperceptible to semantics-unchanging) adversarial perturbations on the input~\cite{szegedy2014iclr}.

Given a learned model $f:X\rightarrow Y$ (e.g. a classifier) and an input $x$, we say that an additive input perturbation $\delta$ is an adversarial perturbation of $x$ for $f$ if $\delta$ is small (e.g. $||\delta||_2<\epsilon$ for some $\epsilon>0$) and the new output $f(x+\delta)$ is significantly different from the original $f(x)$ (e.g. a different class prediction). Omnipresence of such examples throughout the input space against most existing neural network architectures has spurred discussions over the safety of neural network applications in security-critical tasks, such as self-driving cars.

For generating adversarial perturbations, people have mostly considered using diverse variants of the gradient over the input~\cite{goodfellow2015iclr,moosavi2016cvpr,moosavi2016universal,joon17iccv}. The simplest of them is the fast gradient sign method (FGSM,~\cite{goodfellow2015iclr}) which computes the following quantity:
\begin{align}
\delta = -\epsilon\cdot\text{sgn}(\nabla_x f^y(x))
\end{align}
the negative signed input gradient for the prediction of class $y$, the argmax prediction by $f$. While being simple and effective, this requires expensive gradient computation for every input $x$ and is hard to integrate into other learning models.

Baluja et al.~\cite{baluja2017adversarial} have proposed the Adversarial Transformer Network (ATN), which, instead of relying on gradient computations, obtains the perturbations through a learned feed-forward network $g(x)$. The network is learned through the following objective
\begin{align}
\underset{\theta}{\min} \underset{x\sim D}{\mathbb{E}}[f^y(x+g_\theta(x))] \\
\text{s.t. }||g_\theta(x)||_2\leq \epsilon \text{ for all }x \label{eq:constraint}
\end{align}
via stochastic gradient descent over multiple training images $x\sim D$. In our work, we use the ATN as the perturbation generator against the DQN: $Q(x+g_\theta(x))$. We will explain the method for training $g_\theta$ to impose adversarial reward on $Q$ in the next section.

\begin{figure*}[!ht]
	\centering 
	\includegraphics[width=1.5\columnwidth]{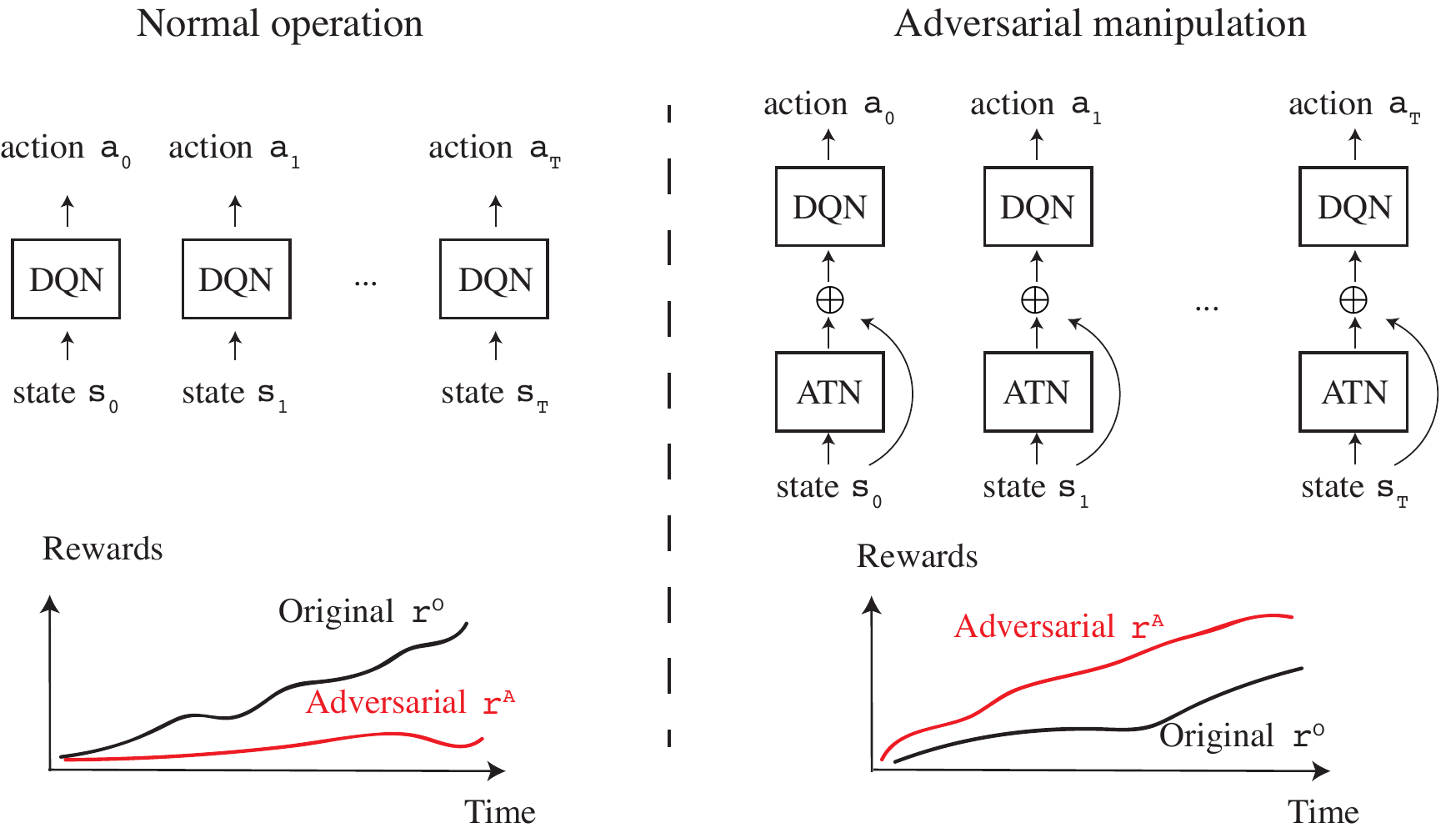}
	\caption{\label{threat-model} Overview of our threat model. We show the victim agent's operation under normal operation (left) and under adversarial manipulation (right). In the threat model, the victim Deep Q-Network (DQN) trained for the original reward $r^O$ is manoeuvred to pursue an adversarial reward $r^A$ as adversarially perturbed sequence of inputs are fed (ATNs). Under this adversarial manipulation, the fooled agent increases the adversarial reward ($r^A$, red curves) over time.}
\end{figure*}
\begin{figure}[!h]
	\centering 
	\includegraphics[width=0.93\columnwidth]{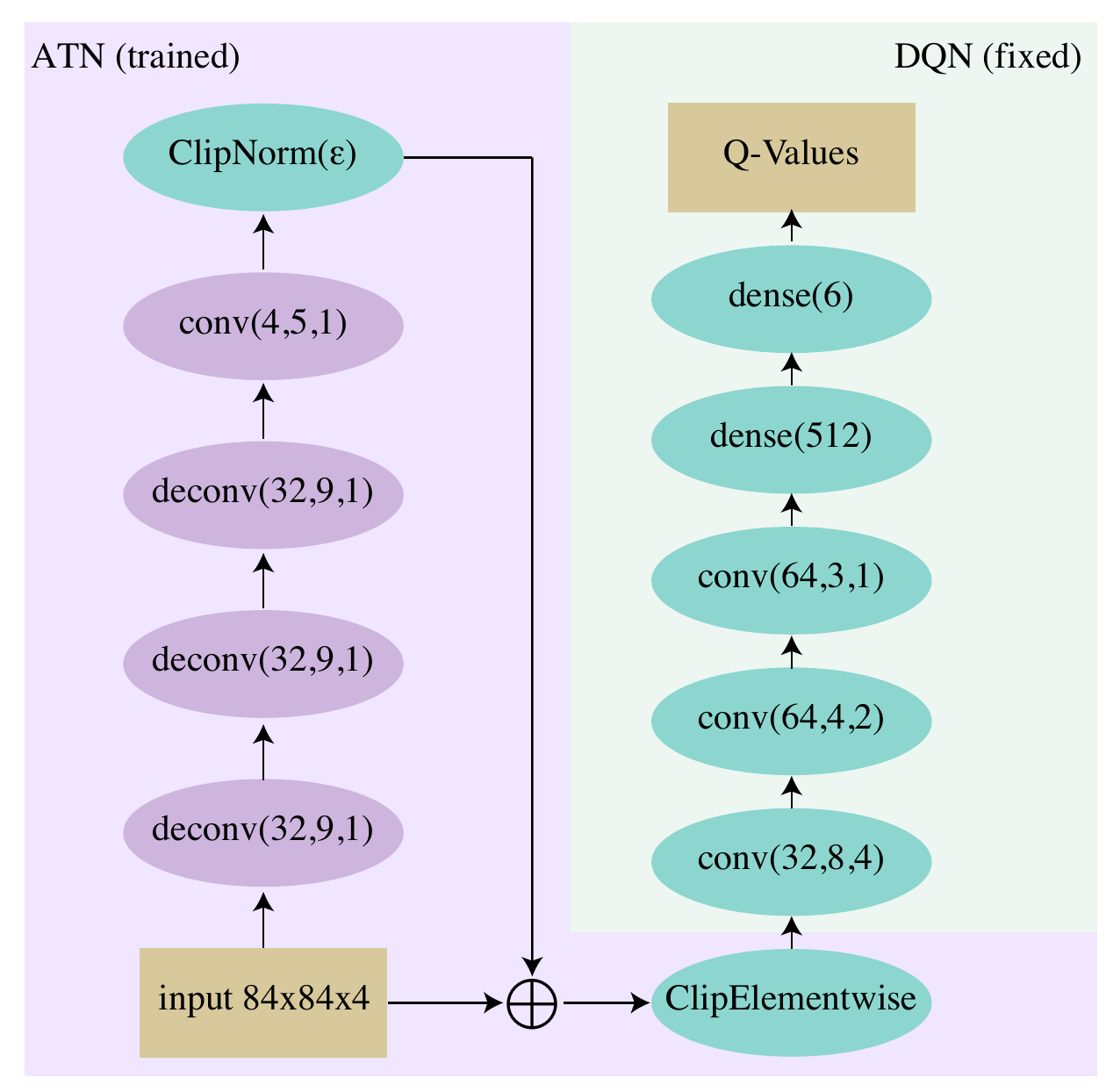}
	\caption{\label{architecture}Architecture of our threat model. ATN takes input frames, computes perturbations in feed-forward fashion, and adds the perturbations back on the input. The perturbed input is fed to the victim DQN. During training, DQN parameters are fixed (green ovals), while ATN parameters are updated (purple ovals). (De-)Convolution parameters indicate (\#\ignorespaces filters, kernel size, stride) and the dense parameter indicates (\#\ignorespaces features). For ATN deconvolution layers, we always use dilation rate 2.
	}
\end{figure}

\section{Threat Model}
\label{sec:4threatmodel}

We consider an adversary whose goal is to make a trained victim agent interacting with an environment for the original reward $r^O$ to maximise an arbitrary adversarial reward $r^A$ through a sequence of state perturbations. An overview of our approach is in Figure~\ref{threat-model}. In this section, we  describe in detail how the perturbations are computed to guide the agent towards the adversarial reward $r^A$, and then discuss key assumptions for our threat model. 

\subsection{Attack Algorithm}

See the right half of Figure \ref{threat-model} for an overview of our attack paradigm. Given a fixed victim policy network $Q_\phi$ trained for the original reward $r^O$, we attach the Adversarial Transformer Network (ATN)~\cite{baluja2017adversarial}, a feedforward deep neural network $g_\theta:X\rightarrow X$ which computes the perturbation to be added to the input of the victim DQN $Q_\phi$. The aim of the adversary is to learn $\theta$ such that the perturbed states lead the victim to follow an arbitrary adversarial reward $r^A$.

We approach the training of $\theta$ by regarding the combination of DQN and ATN $Q_\phi(x+g_\theta(x)))$ as another DQN to be trained for the adversarial reward $r^A$. In this process, we fix the parameters learned for the victim $Q_\phi$ and only learn $\theta$. 
Specifically, we solve for the Equation \ref{eq:DQN-objective} where $Q_\phi$ is now the mapping $x\mapsto Q_\phi(x+g_\theta(x))$, and the trained parameters are $\theta$ (and the victim DQN parameters~$\phi$ are fixed).
Using the generalisability of $g_\theta$ to unseen states $x$, the adversary then only needs to feed the input state through $g_\theta$ and then through the victim DQN to achieve the desired outcome. 

The detailed architecture is shown in Figure~\ref{architecture}. Note that the victim DQN architecture is the same as in Mnih et al.~\cite{mnih2015human}.
To enforce the norm constraint in Equation~\ref{eq:constraint}, we insert a norm-clipping layer (ClipNorm) which does the following operation:
\begin{align}
\text{clip}_p(x)=
\begin{cases}
\frac{p\cdot x}{||x||_2},& \text{if } ||x||_2\geq p\\
x,              & \text{otherwise}
\end{cases}
\label{eq:clipping}
\end{align}
We parametrize $p$ using $\epsilon$ such that $p = 84 \cdot 84 \cdot 4 \cdot \epsilon$.

We also enforce the $[0,1]$ range for the input values to the victim DQN via a ClipElementwise layer: $x\mapsto \min(\max(x,0),1)$.
Spatial dimensions for the intermediate features do not change throughout the ATN network $g_\theta$.

\subsection{Assumptions}

We consider an adversary which can manoeuvre the long-term behaviour and goal for a victim agent only through a sequence of small input perturbations rather than through direct manipulation of the victim's policy network.

We explicitly spell out the assumptions we make for the described algorithm and our experimental evaluations. While some are restrictive, others may be easily relaxed.

\noindent
\textbf{1. White-box access at training time.} For training the ATN, the adversary requires gradient access to the victim policy network. While this is restrictive, there exists much ongoing work on the transferability of adversarial examples. Measuring the efficacy of those techniques for our adversary will be an interesting future work.

\noindent
\textbf{2. Manipulation on the input stream.} We assume that the adversary can manipulate the input stream (state observations) for the victim DQN. This can be achieved e.g. by hacking into  sensors~\cite{kim2017hacking} or by making physical changes to the environment~\cite{kurakin2016adversarial}.

\noindent
\textbf{3. Computational resources to train the ATN.} Training the ATN can be computationally prohibitive for many. However, even one entity with the intention and ability to train such an ATN can be a grave threat in security-critical applications. 

\noindent
\textbf{4. Environment for training the victim.} For training the ATN, we have trained the combined ATN+DQN module on the same environment that has trained the original DQN. This assumption may be relaxed in the future by experimenting with different victim-attacker environments (albeit with the same task).

\noindent
\textbf{5. Fixed victim DQN.} If the victim DQN is updated, then the adversary needs to re-train the ATN. However, in practice, such an update does not occur continually.

\section{Experiments}
\label{sec:5experiments}

\begin{figure*}
	\centering {
		\hfill
		\subfigure[Original reward]{\includegraphics[width=0.8\columnwidth]{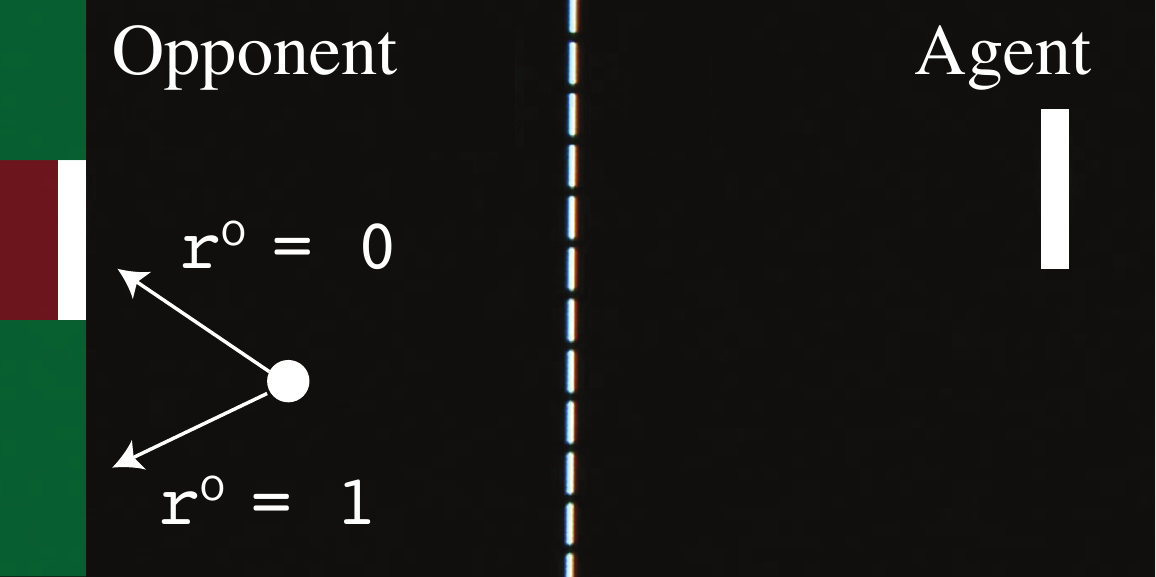}}
		\hfill
		\subfigure[Center reward]{\includegraphics[width=0.8\columnwidth]{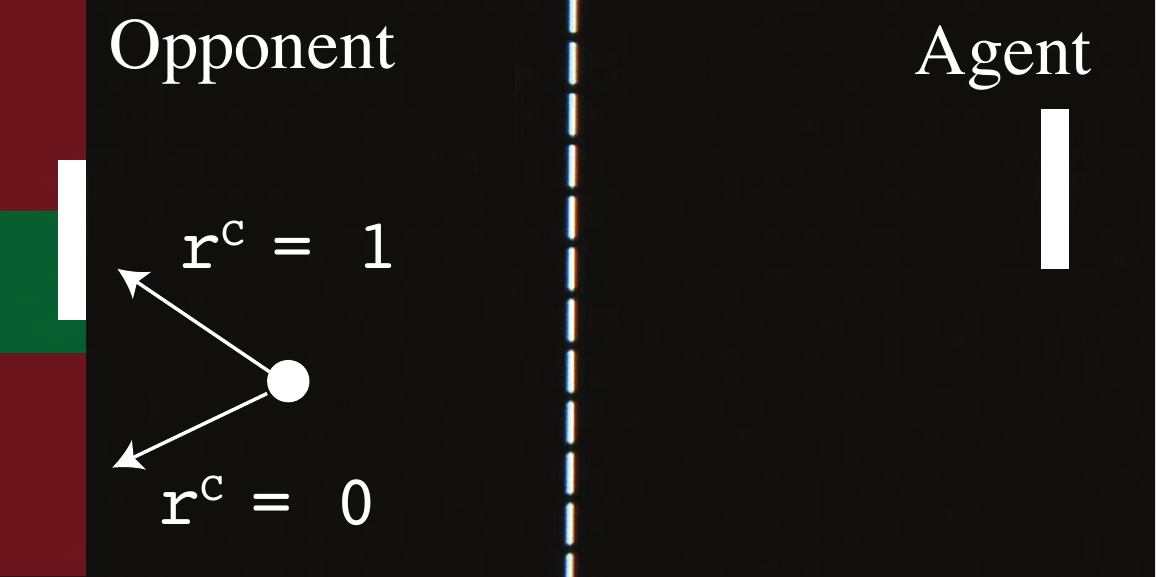}}
		\hfill
	}
	\caption{\label{fig:rewards}Rewards visualised. Green indicates the rewarded regions. Original reward $r^O$ is awarded when the ball does not hit opponent's pad; centre reward $r^C$ is awarded when the ball hits any mid-20\% region, regardless of the opponent's position.} 
\end{figure*}

We evaluate the threat model discussed in the previous section against victim agents trained to play the game of Pong~\cite{openaigym}. We will first describe the game along with the original reward $r^O$ and define an adversarial reward $r^A$ (\S\ref{subsec:5game}). We will discuss implementation details and our evaluation metric in \S\ref{subsec:5details}, and present results and analysis in \S\ref{subsec:5results}.

\subsection{The Game of Pong}
\label{subsec:5game}

In our experiments we focus on the game of Pong~\cite{openaigym}, a classic environment for deep reinforcement learning. It is a great environment for our purposes, since victim agents can be trained to achieve optimal play, a property not enjoyed in more complicated environments. We verify that our attack works even for high-performance victim agents.

We use the Pong simulation from the OpenAI Gym \cite{openaigym}. In this game, two players are positioned on opposite sides of the screen. They can only move up and down. Similar to tennis, a single ball is passed between the two players. The goal of the game is to play the ball such that the other player is unable to catch it. In single-player mode, the opponent uses simple heuristics to play. The original reward $r^O$ is defined as
\begin{align}
r^O(s_t)=
\begin{cases}
1, & \text{if the ball leaves the frame on the opposing side, }\\
-1, & \text{if the ball leaves the frame on the agent's side, }\\
0,              & \text{otherwise.}
\end{cases}
\end{align}

A game state is represented as a $210\times 160$ colour image. Before passing states to the DQN, we apply the same pre-processing as \cite{mnih2015human}. (1) Merge four consecutive frames using a pixel-wise max operation on each channel. (2) Resulting image is converted to grey-scale and down-sampled to $84\times 84$. (3) To enable the DQN to utilise temporal dependencies, four most recent such processed images are put into a queue and then used as inputs to the DQN. (4) To get the next input, we append a new processed frame to the queue and remove the oldest frame. During training, we append about thirty processed frames per second to the queue. To make this pre-processing consistent with the environment, the same action is repeated four times.

The action space consists of six actions available on the Atari controller (joystick): four directions, one button, and the ``no action'' case.

\subsubsection{Victim Agents}

As victim agents, we use three off-the-shelf trained agents from the OpenAI baselines~\cite{openAIbaselines} (\texttt{OAI1}, \texttt{OAI2}, \texttt{OAI3}), as well as five agents trained by us independently (\texttt{OR1}, \texttt{OR2}, \texttt{OR3}, \texttt{OR4}, \texttt{OR5}). They are all trained for the original reward $r^O$ of winning the game.

\subsubsection{Adversarial Reward}
\label{subsec:5advreward}

The adversary intends to impose another reward $r^A$ on the trained agents. In our work, we consider the \emph{centre reward}, $r^C$. For any given time step $t$, the centre reward is defined as
\begin{align}
r^C(s_t)=
\begin{cases}
1, & \text{if the ball hits the centre 20\% of the enemy line, }\\
0,              & \text{otherwise.}
\end{cases}
\end{align}
See Figure~\ref{fig:rewards} for an illustration of the rewards.

\subsubsection{FGSM Baseline}
As a baseline, we adapt the work of Behzadan et al.~\cite{BehzadanM17} to our setting. An FGSM adversary is inserted in between the victim agent and its input. Since we assume to have white-box access at training time of the ATN, we grant the FGSM adversary white-box access to the victim agent. Instead of an ATN, the FGSM adversary consists of two components: a policy DQN and a perturbation generation module. The policy DQN determines the desired action of the FGSM adversary. We then generate a perturbation using FGSM on the victim agent, where the desired action of the policy DQN defines the one-hot target distribution. This perturbation is then added to the input, clipped element-wise, and given to the victim agent network.

\subsection{Implementation Details}
\label{subsec:5details}

We describe the training procedure for the victim DQN agents as well as the adversary's ATN. 

\subsubsection{Training Victim Agents}

Here, we describe the training details for the victim agents mentioned above (\texttt{OAI1}-\texttt{OAI3} and \texttt{OR1}-\texttt{OR5}).
There are two differences in the training of the \texttt{OAI} agents and the \texttt{OR} agents. The \texttt{OAI} agents and the \texttt{OR} agents were trained on 200 million frames and 80 million frames, respectively.\footnote{In Pong, agents already achieve optimal reward after about 50 million frames.} We use a different schedule for the fraction of random actions: during training of the \texttt{OR} agents, we linearly anneal from $1.0$ to $0.1$ for 8 million frames, and then keep it constant at $0.1$ afterwards. For the \texttt{OAI} agents, the rate is linearly annealed from $1.0$ to $0.1$ for 4 million frames, and then linearly annealed to $0.01$ for another 36 million frames, where it is then kept constant.

The DQN replay buffer size has been set to 100k observation tuples. We use prioritized replay~\cite{schaul2015prioritized}. We use Adam with a learning rate of \num{1e-4} and $\gamma=0.99$~\cite{kingma2014adam}, with the batch size 32. Every agent uses a different random seed to initialize the network parameters and the environment.

\subsubsection{Training Adversary}

Given eight victim agents, we train eight corresponding Adversarial Transformer Networks (ATNs) for the centre reward $r^C$. The training procedure for the composite ATN+DQN network is identical to the training of victims, except for different random seeds and a different schedule for the fraction of random actions taken (linearly anneal from $1.0$ to $0.1$ for 40 million frames, then keep it constant).
We control the amount of perturbation via the norm clipping layer (Equation~\ref{eq:clipping}). Unless otherwise denoted, we use $\epsilon=10^{-4}$.

To measure the oracle performance of ATNs, we have trained five additional vanilla DQNs from scratch for the centre reward: \texttt{CR1}-\texttt{CR5}.

\subsubsection{Evaluation Metric}

We evaluate the victim agents' performance on the original and adversarial rewards with or without the adversarial manipulation on the input stream. The victim agents play the game of Pong from five different seeds for $40$k frames each, without taking any random actions. We then plot the average accumulative rewards over the five random seeds.

\subsubsection{FGSM Baseline}
We consider FGSM adversaries with \texttt{CR1-CR5} as policy networks. We use all of the agents trained for $r^O$, \texttt{OAI1-OAI3} and \texttt{OR1-OR5}, as victim agents. In a grid search, we empirically determined the FGSM perturbation norm that yields the highest success rate in imposing the policy DQN's desired action on the victim agents. This norm is $\epsilon = 1\times 10^{-5}$. The average success rate varies from $49\%$ to $61\%$ for the \texttt{OAI} victim agents, and from $86\%$ to $95\%$ for the \texttt{OR} victim agents.

\subsection{Results}
\label{subsec:5results}

\begin{figure*}
\centering {
\hfill
\subfigure[\label{subfig:main-result-original}Original reward $r^O$]{\includegraphics[width=0.3\textwidth]{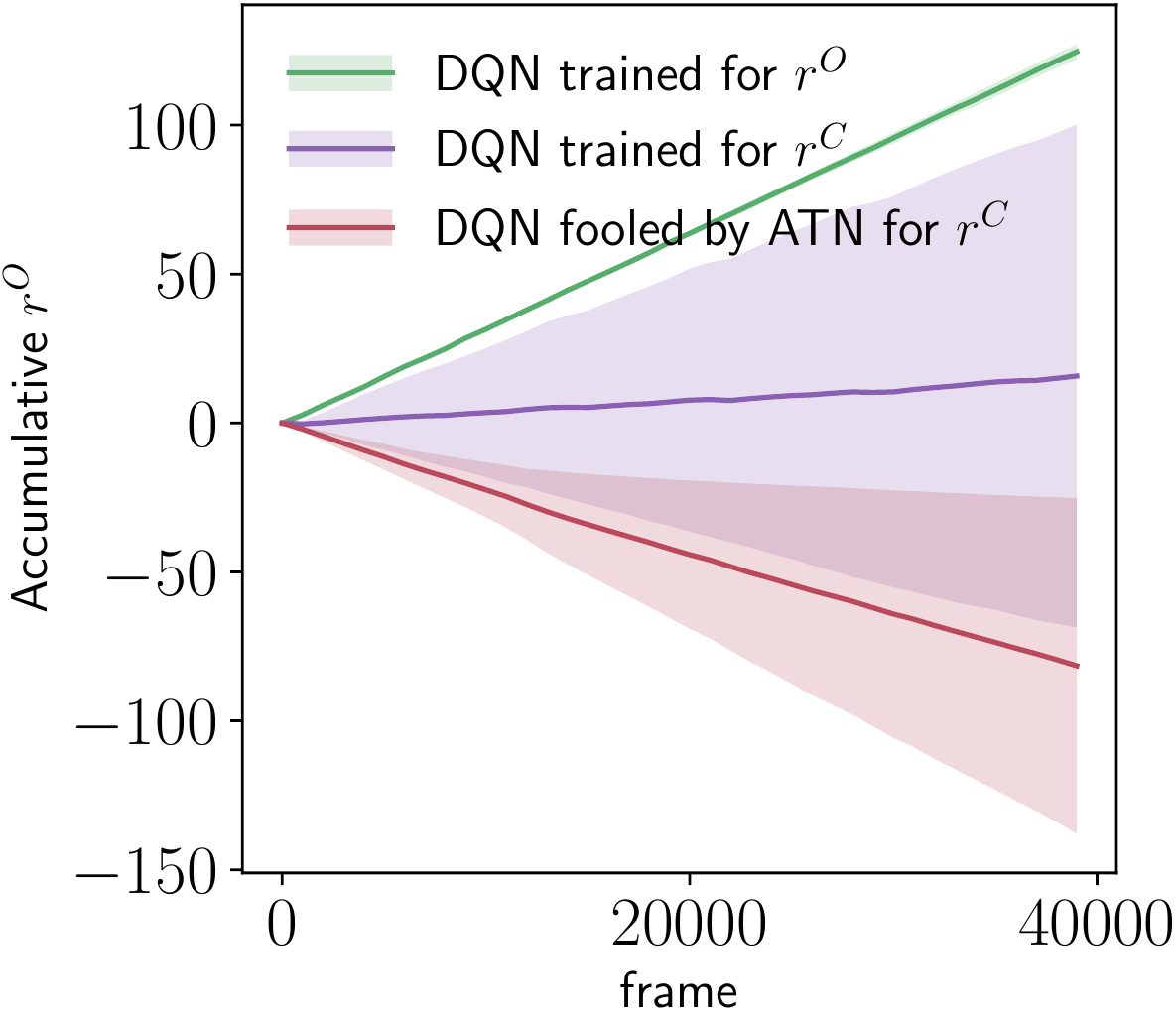}}
\hfill
\subfigure[\label{subfig:main-result-centre}Centre reward $r^C$ reward]{\includegraphics[width=0.3\textwidth]{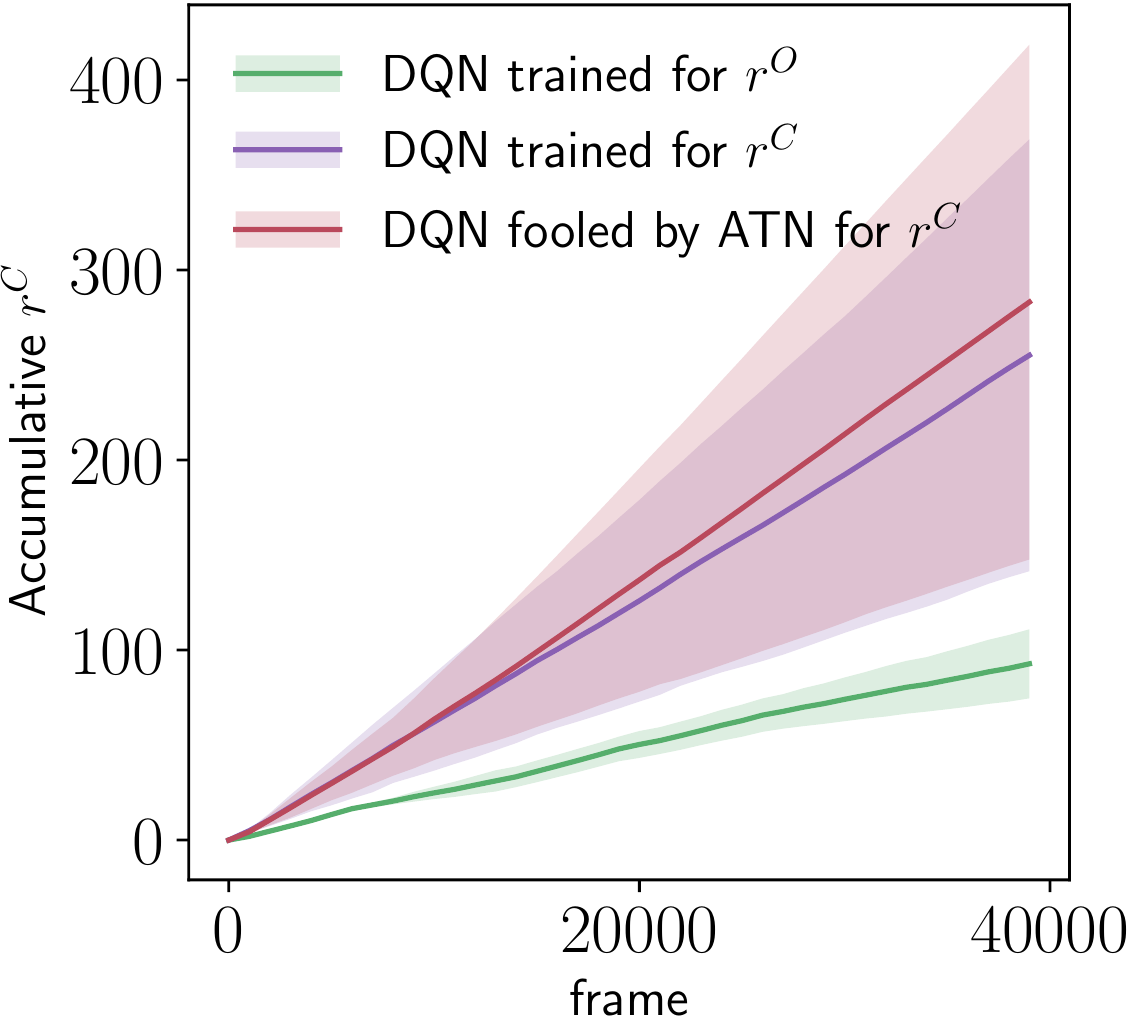}}
\hfill
}
\caption{\label{fig:main-result}Accumulative rewards for agents (1) trained for original reward $r^O$ (2) trained for centre reward $r^C$, and (3) trained for $r^O$ but  manipulated towards $r^C$ by our ATN adversary.  Curves are averaged over multiple independently trained agents; error bands indicate $\pm 1$ standard deviation.} 
\end{figure*}
\begin{figure*}
\centering {
\hfill
\subfigure[\label{subfig:baseline-avg}Averaged over agents]{\includegraphics[width=0.3\textwidth]{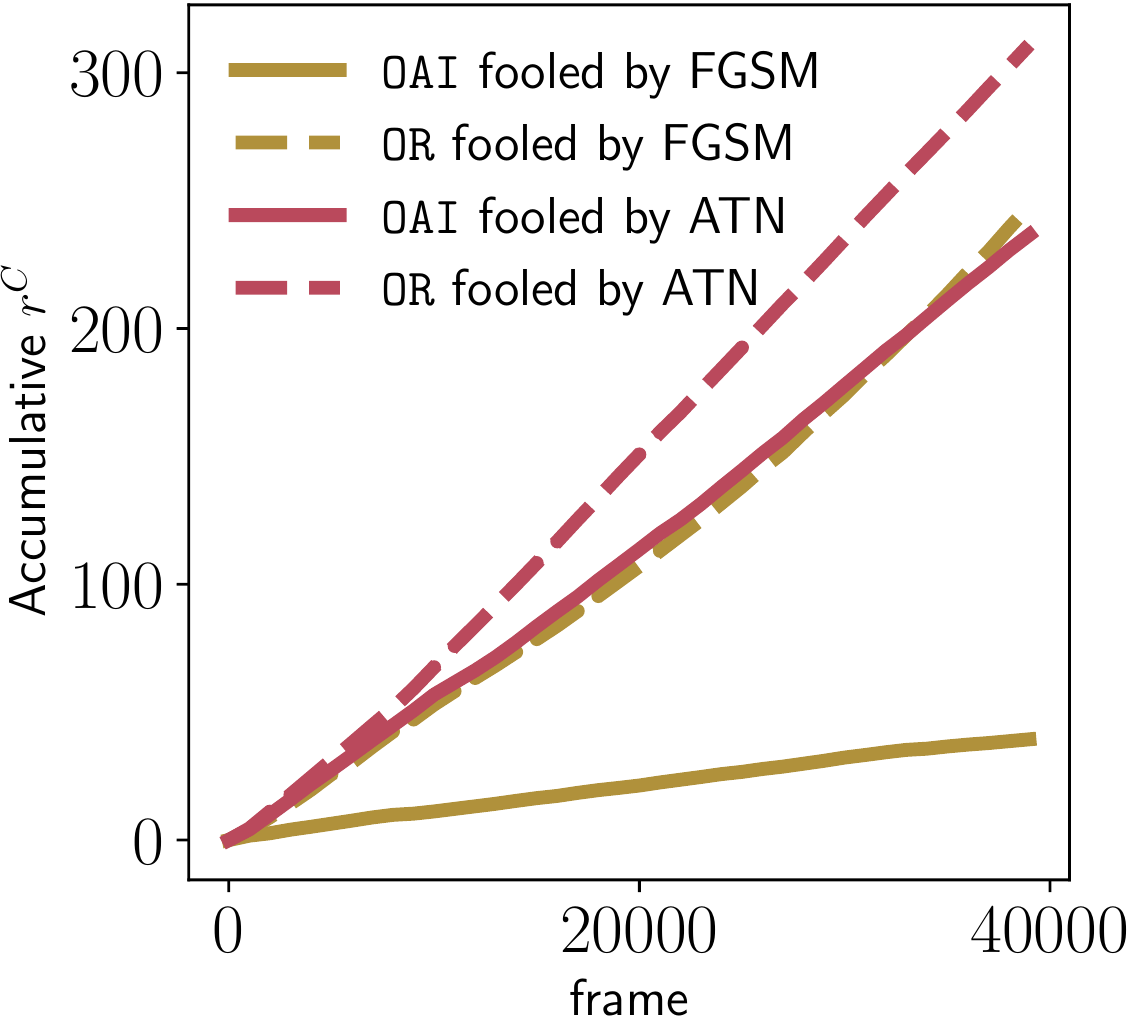}}
\hfill
\subfigure[\label{subfig:baseline-all}All agents]{\includegraphics[width=0.3\textwidth]{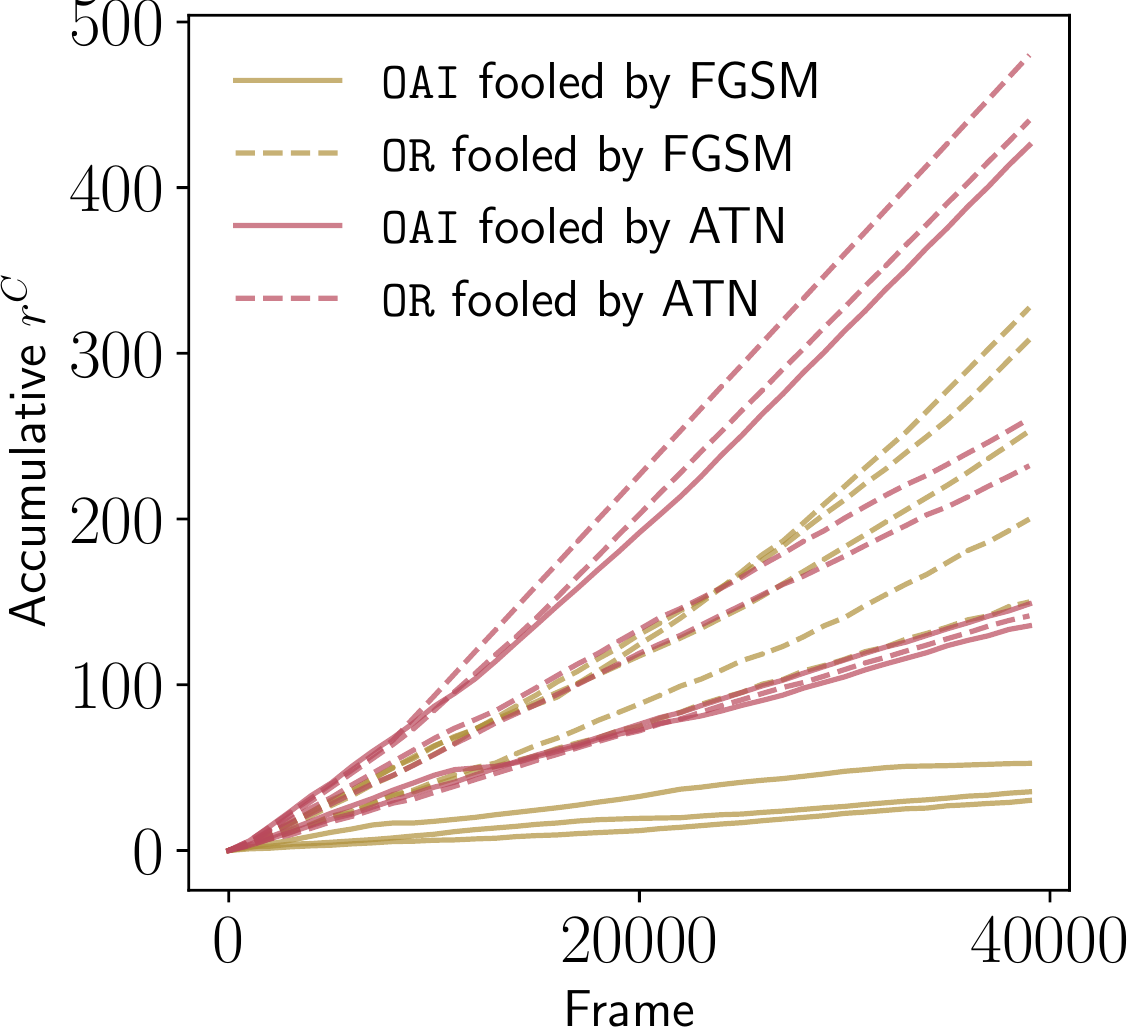}}
\hfill
}
\caption{\label{fig:baseline-comparison}Accumulative centre rewards $r^C$ for two types of agents (\texttt{OAI}: OpenAI pretrained and \texttt{OR}: our trained models) each fooled into $r^C$ by two methods (FGSM baseline and our ATN based adversarial policy enforcement).} 
\end{figure*}

\begin{figure}
	\centering 
	\includegraphics[width=0.25\textwidth]{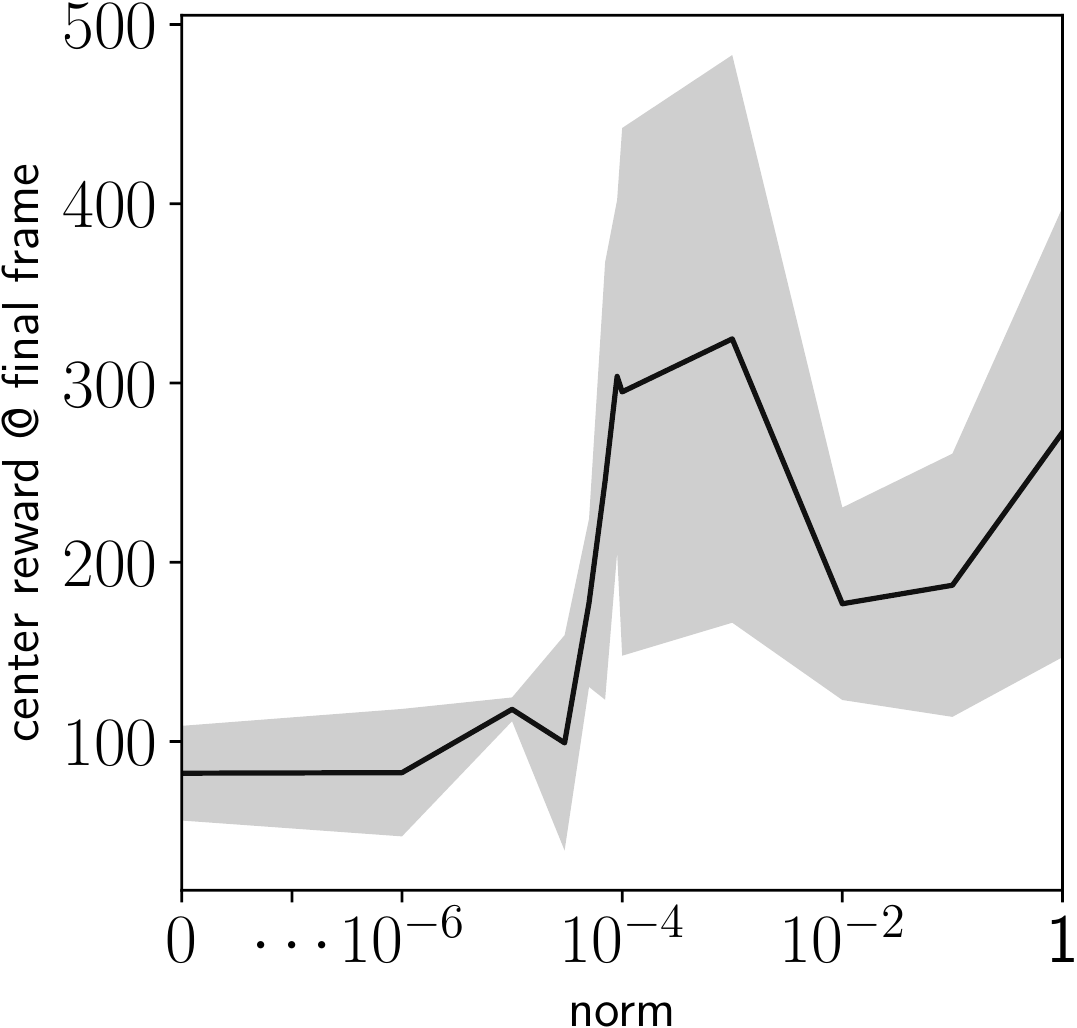}
	\caption{\label{fig:norm}Final frame ($40\,000^\text{th}$ frame) center reward for agents trained for original reward but adversarially guided by the center reward, when the adversary is given different amounts of $L_2$ norm budgets (x axis). Error band indicates $\pm 1$ standard deviation.} 
\end{figure}

We present experimental results here. See Figures~\ref{fig:main-result}, \ref{fig:norm}, and \ref{fig:perturbation_visualisation}.

\subsubsection{Main Results}

We first examine the performance of the eight victim agents (\texttt{OAI1}-\texttt{OAI3} and \texttt{OR1}-\texttt{OR5}) for the original reward $r^O$ of winning the games. In Figure~\ref{subfig:main-result-original}, we observe that these victim agents accrue $r^O$ steadily over time, all reaching over 100 rewarded points at the final frame (40k). We confirm that the victim agents are fully performant at playing the game.

In terms of the centre reward $r^C$, the eight victim agents achieve around 100 rewarded points at the final frame (Figure~\ref{subfig:main-result-centre}). Although they are not explicitly trained for $r^C$, they sometimes send the ball to the middle while trying to win the games.

We then study the performance of the vanilla DQN networks trained from scratch for the centre reward $r^C$, to determine if the task is learnable at all. In Figure~\ref{subfig:main-result-centre}, we confirm that the DQN trained for $r^C$ attains a far better performance at accruing $r^C$ than do the eight victim agents. We note that agents trained for $r^C$ are not doing well for the original task of winning the games ($r^O$). We confirm that it is possible to build a policy towards $r^C$.

Finally, we examine the ability of our adversary to generate perturbations that misguide agents to aim for the alternative reward $r^C$. In Figure~\ref{subfig:main-result-centre}, we observe that the victim DQNs fooled into pursuing $r^C$ do effectively accrue $r^C$ over time, matching the performance of DQNs trained from scratch to pursue $r^C$. Our adversary can successfully impose an adversarial policy on a victim agent through a sequence of perturbations.

\subsubsection{$L_2$ Norm Restriction}

In the previous set of experiments, we have used the $L_2$ norm constraint $\epsilon=10^{-4}$. Here, we study the effect of the norm constraint on the effectiveness of attacks. It is expected that relaxing the norm constraints gives more freedom for the adversary to choose the adversarial patterns that effectively lead the victim to the adversarial goal.

See Figure~\ref{fig:norm} for the results. We show the final frame centre rewards versus $\epsilon$ for eight vanilla agents each fooled by our sequential adversarial attacks. We indeed observe that, from $\epsilon=0$ (no attack) to $\epsilon=10^{-5}$, the final frame $r^C$ increases, confirming that the adversary is better-off with relaxed norm constraints. However, for $\epsilon>10^{-4}$, greater variances in the final rewards are observed; we conjecture that for such great amount of perturbations training is unstable and does not converge.

\begin{figure*}[h!]%
	\centering
	\setlength{\tabcolsep}{0.3em}
	\newcommand{\imwidth}{0.23}
	\begin{tabular}{ccccc}
		$\varepsilon$ && Unperturbed & Perturbation & Perturbed \\
		\vspace{0em} & \\
		{\rotatebox{90}{\hspace{4em}{$1\times 10^{-5}$}\hspace{0em}}} &&
		\includegraphics[width=\imwidth\textwidth]{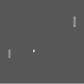} & 
		\includegraphics[width=\imwidth\textwidth]{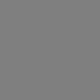} & 
		\includegraphics[width=\imwidth\textwidth]{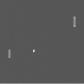} \\
		{\rotatebox{90}{\hspace{4em}{$3\times 10^{-5}$}\hspace{0em}}} &&
		\includegraphics[width=\imwidth\textwidth]{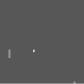} & 
		\includegraphics[width=\imwidth\textwidth]{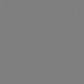} & 
		\includegraphics[width=\imwidth\textwidth]{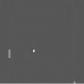} \\
		{\rotatebox{90}{\hspace{4em}{$7\times 10^{-5}$}\hspace{0em}}} &&
		\includegraphics[width=\imwidth\textwidth]{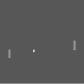} & 
		\includegraphics[width=\imwidth\textwidth]{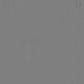} & 
		\includegraphics[width=\imwidth\textwidth]{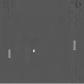} \\
		{\rotatebox{90}{\hspace{4em}{$1\times 10^{-4}$}\hspace{0em}}} &&
		\includegraphics[width=\imwidth\textwidth]{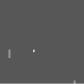} & 
		\includegraphics[width=\imwidth\textwidth]{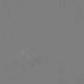} & 
		\includegraphics[width=\imwidth\textwidth]{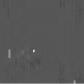} \\
	\end{tabular}
	\caption{Examples of adversarial perturbations at $L_2$ norm constraints $\varepsilon\in\{1\times 10^{-5},3\times 10^{-5},7\times 10^{-5},1\times 10^{-4}\}$ that guides the agents to pursue the centre reward.}%
	\label{fig:perturbation_visualisation}%
\end{figure*}

We visualise the amount of perturbations at each $\epsilon$ level in Figure~\ref{fig:perturbation_visualisation}. Note that $\epsilon=10^{-4}$ make the perturbations visible, but this would not interfere with a human player.

\subsubsection{FGSM Baseline}
We now compare our method to the FGSM adversary. In Figure~\ref{subfig:baseline-all}, we see that, averaged over the five policy networks \texttt{CR1-CR5}, the FGSM adversary achieves a significantly lower accumulated centre reward at the final frame for the \texttt{OAI} agents than for the \texttt{OR} agents. The latter performance is about on par with the ATN adversary and the agents trained for the centre reward, \texttt{CR1-CR5}. When considering only the \texttt{OAI} agents, Figure~\ref{subfig:baseline-avg} shows that the ATN adversary outperforms the FGSM adversary. We hypothesize that the longer training and the different schedule for the random exploration causes the \texttt{OAI} agents to become more robust to the FGSM adversary. This is supported by the significantly higher success rate by the FGSM adversary in imposing its desired action on the \texttt{OR} agents compared to the \texttt{OAI} agents. Possibly due to its more intimate joint training with the victim agent, the ATN adversary can still successfully attack the \texttt{OAI} agents.

\section{Conclusion}
\label{sec:7conclusion}

We have exposed a new security threat for deeply learned policies. Much prior work has argued that a small perturbation on the states can make a deeply learned policy fail to achieve the originally set task. In our work, we have shown experimentally that it is moreover possible to impose an arbitrary adversarial reward and corresponding policy on a policy network (Deep Q-Network) through a sequence of perturbations on the input stream (state observations). The possibility of such an adversary questions the safety of deploying learned agents in everyday applications, not to mention security-critical ones.

\clearpage
\bibliographystyle{ACM-Reference-Format}
\bibliography{bib}


\begin{thebibliography}{38}


\ifx \showCODEN    \undefined \def \showCODEN     #1{\unskip}     \fi
\ifx \showDOI      \undefined \def \showDOI       #1{#1}\fi
\ifx \showISBNx    \undefined \def \showISBNx     #1{\unskip}     \fi
\ifx \showISBNxiii \undefined \def \showISBNxiii  #1{\unskip}     \fi
\ifx \showISSN     \undefined \def \showISSN      #1{\unskip}     \fi
\ifx \showLCCN     \undefined \def \showLCCN      #1{\unskip}     \fi
\ifx \shownote     \undefined \def \shownote      #1{#1}          \fi
\ifx \showarticletitle \undefined \def \showarticletitle #1{#1}   \fi
\ifx \showURL      \undefined \def \showURL       {\relax}        \fi
\providecommand\bibfield[2]{#2}
\providecommand\bibinfo[2]{#2}
\providecommand\natexlab[1]{#1}
\providecommand\showeprint[2][]{arXiv:#2}

\bibitem[\protect\citeauthoryear{Baluja and Fischer}{Baluja and
  Fischer}{2018}]%
        {baluja2017adversarial}
\bibfield{author}{\bibinfo{person}{Shumeet Baluja} {and} \bibinfo{person}{Ian
  Fischer}.} \bibinfo{year}{2018}\natexlab{}.
\newblock \showarticletitle{Learning to Attack: Adversarial Transformation
  Networks}. In \bibinfo{booktitle}{\emph{Proceedings of AAAI-2018}}.
\newblock
\urldef\tempurl%
\url{http://www.esprockets.com/papers/aaai2018.pdf}
\showURL{%
\tempurl}


\bibitem[\protect\citeauthoryear{Behzadan and Munir}{Behzadan and
  Munir}{2017}]%
        {BehzadanM17}
\bibfield{author}{\bibinfo{person}{Vahid Behzadan} {and}
  \bibinfo{person}{Arslan Munir}.} \bibinfo{year}{2017}\natexlab{}.
\newblock \showarticletitle{Vulnerability of deep reinforcement learning to
  policy induction attacks}. In \bibinfo{booktitle}{\emph{International
  Conference on Machine Learning and Data Mining in Pattern Recognition}}.
  Springer, \bibinfo{pages}{262--275}.
\newblock


\bibitem[\protect\citeauthoryear{Bellman}{Bellman}{1957}]%
        {bellman1957markovian}
\bibfield{author}{\bibinfo{person}{Richard Bellman}.}
  \bibinfo{year}{1957}\natexlab{}.
\newblock \showarticletitle{A Markovian decision process}.
\newblock \bibinfo{journal}{\emph{Journal of Mathematics and Mechanics}}
  (\bibinfo{year}{1957}), \bibinfo{pages}{679--684}.
\newblock


\bibitem[\protect\citeauthoryear{Bertsekas, Bertsekas, Bertsekas, and
  Bertsekas}{Bertsekas et~al\mbox{.}}{1995}]%
        {bertsekas1995dynamic}
\bibfield{author}{\bibinfo{person}{Dimitri~P Bertsekas},
  \bibinfo{person}{Dimitri~P Bertsekas}, \bibinfo{person}{Dimitri~P Bertsekas},
  {and} \bibinfo{person}{Dimitri~P Bertsekas}.}
  \bibinfo{year}{1995}\natexlab{}.
\newblock \bibinfo{booktitle}{\emph{Dynamic programming and optimal control}}.
  Vol.~\bibinfo{volume}{1}.
\newblock \bibinfo{publisher}{Athena scientific Belmont, MA}.
\newblock


\bibitem[\protect\citeauthoryear{Brockman, Cheung, Pettersson, Schneider,
  Schulman, Tang, and Zaremba}{Brockman et~al\mbox{.}}{2016}]%
        {openaigym}
\bibfield{author}{\bibinfo{person}{Greg Brockman}, \bibinfo{person}{Vicki
  Cheung}, \bibinfo{person}{Ludwig Pettersson}, \bibinfo{person}{Jonas
  Schneider}, \bibinfo{person}{John Schulman}, \bibinfo{person}{Jie Tang},
  {and} \bibinfo{person}{Wojciech Zaremba}.} \bibinfo{year}{2016}\natexlab{}.
\newblock \bibinfo{title}{Open{AI} {G}ym}.
\newblock
\newblock
\showeprint{arXiv:1606.01540}


\bibitem[\protect\citeauthoryear{Dhariwal, Hesse, Klimov, Nichol, Plappert,
  Radford, Schulman, Sidor, and Wu}{Dhariwal et~al\mbox{.}}{2017}]%
        {openAIbaselines}
\bibfield{author}{\bibinfo{person}{Prafulla Dhariwal},
  \bibinfo{person}{Christopher Hesse}, \bibinfo{person}{Oleg Klimov},
  \bibinfo{person}{Alex Nichol}, \bibinfo{person}{Matthias Plappert},
  \bibinfo{person}{Alec Radford}, \bibinfo{person}{John Schulman},
  \bibinfo{person}{Szymon Sidor}, {and} \bibinfo{person}{Yuhuai Wu}.}
  \bibinfo{year}{2017}\natexlab{}.
\newblock \bibinfo{title}{Open{AI} Baselines}.
\newblock \bibinfo{howpublished}{\url{https://github.com/openai/baselines}}.
\newblock


\bibitem[\protect\citeauthoryear{Dong, Liao, Pang, Hu, and Zhu}{Dong
  et~al\mbox{.}}{2017}]%
        {dong2017discovering}
\bibfield{author}{\bibinfo{person}{Yinpeng Dong}, \bibinfo{person}{Fangzhou
  Liao}, \bibinfo{person}{Tianyu Pang}, \bibinfo{person}{Xiaolin Hu}, {and}
  \bibinfo{person}{Jun Zhu}.} \bibinfo{year}{2017}\natexlab{}.
\newblock \showarticletitle{Discovering Adversarial Examples with Momentum}.
\newblock \bibinfo{journal}{\emph{arXiv preprint arXiv:1710.06081}}
  (\bibinfo{year}{2017}).
\newblock


\bibitem[\protect\citeauthoryear{Gandhi, Pinto, and Gupta}{Gandhi
  et~al\mbox{.}}{2017}]%
        {gandhi2017learning}
\bibfield{author}{\bibinfo{person}{Dhiraj Gandhi}, \bibinfo{person}{Lerrel
  Pinto}, {and} \bibinfo{person}{Abhinav Gupta}.}
  \bibinfo{year}{2017}\natexlab{}.
\newblock \showarticletitle{Learning to fly by crashing}.
\newblock \bibinfo{journal}{\emph{arXiv preprint arXiv:1704.05588}}
  (\bibinfo{year}{2017}).
\newblock


\bibitem[\protect\citeauthoryear{Goodfellow, Shlens, and Szegedy}{Goodfellow
  et~al\mbox{.}}{2015}]%
        {goodfellow2015iclr}
\bibfield{author}{\bibinfo{person}{Ian~J Goodfellow}, \bibinfo{person}{Jonathon
  Shlens}, {and} \bibinfo{person}{Christian Szegedy}.}
  \bibinfo{year}{2015}\natexlab{}.
\newblock \showarticletitle{Explaining and harnessing adversarial examples}. In
  \bibinfo{booktitle}{\emph{ICLR}}.
\newblock


\bibitem[\protect\citeauthoryear{Huang, Papernot, Goodfellow, Duan, and
  Abbeel}{Huang et~al\mbox{.}}{2017}]%
        {HuangPGDA17}
\bibfield{author}{\bibinfo{person}{Sandy~H. Huang}, \bibinfo{person}{Nicolas
  Papernot}, \bibinfo{person}{Ian~J. Goodfellow}, \bibinfo{person}{Yan Duan},
  {and} \bibinfo{person}{Pieter Abbeel}.} \bibinfo{year}{2017}\natexlab{}.
\newblock \showarticletitle{Adversarial Attacks on Neural Network Policies}.
\newblock \bibinfo{journal}{\emph{CoRR}}  \bibinfo{volume}{abs/1702.02284}
  (\bibinfo{year}{2017}).
\newblock
\showeprint[arxiv]{1702.02284}
\urldef\tempurl%
\url{http://arxiv.org/abs/1702.02284}
\showURL{%
\tempurl}


\bibitem[\protect\citeauthoryear{Kim}{Kim}{2017}]%
        {kim2017hacking}
\bibfield{author}{\bibinfo{person}{Yongdae Kim}.}
  \bibinfo{year}{2017}\natexlab{}.
\newblock \showarticletitle{Hacking Sensors}.
\newblock  (\bibinfo{year}{2017}).
\newblock


\bibitem[\protect\citeauthoryear{Kingma and Ba}{Kingma and Ba}{2014}]%
        {kingma2014adam}
\bibfield{author}{\bibinfo{person}{Diederik~P Kingma} {and}
  \bibinfo{person}{Jimmy Ba}.} \bibinfo{year}{2014}\natexlab{}.
\newblock \showarticletitle{Adam: A method for stochastic optimization}.
\newblock \bibinfo{journal}{\emph{arXiv preprint arXiv:1412.6980}}
  (\bibinfo{year}{2014}).
\newblock


\bibitem[\protect\citeauthoryear{Ko{\l}cz and Teo}{Ko{\l}cz and Teo}{2009}]%
        {kolcz2009feature}
\bibfield{author}{\bibinfo{person}{Aleksander Ko{\l}cz} {and}
  \bibinfo{person}{Choon~Hui Teo}.} \bibinfo{year}{2009}\natexlab{}.
\newblock \showarticletitle{Feature weighting for improved classifier
  robustness}. In \bibinfo{booktitle}{\emph{CEAS’09: sixth conference on
  email and anti-spam}}.
\newblock


\bibitem[\protect\citeauthoryear{Kos, Fischer, and Song}{Kos
  et~al\mbox{.}}{2017}]%
        {kos2017adversarial}
\bibfield{author}{\bibinfo{person}{Jernej Kos}, \bibinfo{person}{Ian Fischer},
  {and} \bibinfo{person}{Dawn Song}.} \bibinfo{year}{2017}\natexlab{}.
\newblock \showarticletitle{Adversarial examples for generative models}.
\newblock \bibinfo{journal}{\emph{arXiv preprint arXiv:1702.06832}}
  (\bibinfo{year}{2017}).
\newblock


\bibitem[\protect\citeauthoryear{Kos and Song}{Kos and Song}{2017}]%
        {kos2017delving}
\bibfield{author}{\bibinfo{person}{Jernej Kos} {and} \bibinfo{person}{Dawn
  Song}.} \bibinfo{year}{2017}\natexlab{}.
\newblock \showarticletitle{Delving into adversarial attacks on deep policies}.
\newblock  (\bibinfo{year}{2017}).
\newblock


\bibitem[\protect\citeauthoryear{Krizhevsky, Sutskever, and Hinton}{Krizhevsky
  et~al\mbox{.}}{2012}]%
        {krizhevsky2012imagenet}
\bibfield{author}{\bibinfo{person}{Alex Krizhevsky}, \bibinfo{person}{Ilya
  Sutskever}, {and} \bibinfo{person}{Geoffrey~E Hinton}.}
  \bibinfo{year}{2012}\natexlab{}.
\newblock \showarticletitle{Imagenet classification with deep convolutional
  neural networks}. In \bibinfo{booktitle}{\emph{Advances in neural information
  processing systems}}. \bibinfo{pages}{1097--1105}.
\newblock


\bibitem[\protect\citeauthoryear{Kurakin, Goodfellow, and Bengio}{Kurakin
  et~al\mbox{.}}{2016}]%
        {kurakin2016adversarial}
\bibfield{author}{\bibinfo{person}{Alexey Kurakin}, \bibinfo{person}{Ian
  Goodfellow}, {and} \bibinfo{person}{Samy Bengio}.}
  \bibinfo{year}{2016}\natexlab{}.
\newblock \showarticletitle{Adversarial examples in the physical world}.
\newblock \bibinfo{journal}{\emph{CoRR,abs/1607.02533}} (\bibinfo{year}{2016}).
\newblock


\bibitem[\protect\citeauthoryear{Lanckriet, Ghaoui, Bhattacharyya, and
  Jordan}{Lanckriet et~al\mbox{.}}{2003}]%
        {Lanckriet2003minimax}
\bibfield{author}{\bibinfo{person}{Gert~R.G. Lanckriet},
  \bibinfo{person}{Laurent~El Ghaoui}, \bibinfo{person}{Chiranjib
  Bhattacharyya}, {and} \bibinfo{person}{Michael~I. Jordan}.}
  \bibinfo{year}{2003}\natexlab{}.
\newblock \showarticletitle{A Robust Minimax Approach to Classification}.
\newblock \bibinfo{journal}{\emph{JLMR}} (\bibinfo{year}{2003}).
\newblock


\bibitem[\protect\citeauthoryear{Lin, Hong, Liao, Shih, Liu, and Sun}{Lin
  et~al\mbox{.}}{[n. d.]}]%
        {LinHLSLS17}
\bibfield{author}{\bibinfo{person}{Yen-Chen Lin}, \bibinfo{person}{Zhang-Wei
  Hong}, \bibinfo{person}{Yuan-Hong Liao}, \bibinfo{person}{Meng-Li Shih},
  \bibinfo{person}{Ming-Yu Liu}, {and} \bibinfo{person}{Min Sun}.}
  \bibinfo{year}{[n. d.]}\natexlab{}.
\newblock \showarticletitle{Tactics of adversarial attack on deep reinforcement
  learning agents}. In \bibinfo{booktitle}{\emph{Proceedings of the 26th
  International Joint Conference on Artificial Intelligence}}.
\newblock


\bibitem[\protect\citeauthoryear{Lin, Liu, Sun, and Huang}{Lin
  et~al\mbox{.}}{2017}]%
        {lin2017detecting}
\bibfield{author}{\bibinfo{person}{Yen-Chen Lin}, \bibinfo{person}{Ming-Yu
  Liu}, \bibinfo{person}{Min Sun}, {and} \bibinfo{person}{Jia-Bin Huang}.}
  \bibinfo{year}{2017}\natexlab{}.
\newblock \showarticletitle{Detecting Adversarial Attacks on Neural Network
  Policies with Visual Foresight}.
\newblock \bibinfo{journal}{\emph{arXiv preprint arXiv:1710.00814}}
  (\bibinfo{year}{2017}).
\newblock


\bibitem[\protect\citeauthoryear{Lowd and Meek}{Lowd and Meek}{2005}]%
        {lowd2005adversarial}
\bibfield{author}{\bibinfo{person}{Daniel Lowd} {and}
  \bibinfo{person}{Christopher Meek}.} \bibinfo{year}{2005}\natexlab{}.
\newblock \showarticletitle{Adversarial learning}. In
  \bibinfo{booktitle}{\emph{Proceedings of the eleventh ACM SIGKDD
  international conference on Knowledge discovery in data mining}}. ACM,
  \bibinfo{pages}{641--647}.
\newblock


\bibitem[\protect\citeauthoryear{Mandlekar, Zhu, Garg, Fei-Fei, and
  Savarese}{Mandlekar et~al\mbox{.}}{2017}]%
        {mandlekar2017adversarially}
\bibfield{author}{\bibinfo{person}{Ajay Mandlekar}, \bibinfo{person}{Yuke Zhu},
  \bibinfo{person}{Animesh Garg}, \bibinfo{person}{Li Fei-Fei}, {and}
  \bibinfo{person}{Silvio Savarese}.} \bibinfo{year}{2017}\natexlab{}.
\newblock \showarticletitle{Adversarially robust policy learning: Active
  construction of physically-plausible perturbations}. In
  \bibinfo{booktitle}{\emph{IEEE International Conference on Intelligent Robots
  and Systems (to appear)}}.
\newblock


\bibitem[\protect\citeauthoryear{Mnih, Kavukcuoglu, Silver, Rusu, Veness,
  Bellemare, Graves, Riedmiller, Fidjeland, Ostrovski, et~al\mbox{.}}{Mnih
  et~al\mbox{.}}{2015}]%
        {mnih2015human}
\bibfield{author}{\bibinfo{person}{Volodymyr Mnih}, \bibinfo{person}{Koray
  Kavukcuoglu}, \bibinfo{person}{David Silver}, \bibinfo{person}{Andrei~A
  Rusu}, \bibinfo{person}{Joel Veness}, \bibinfo{person}{Marc~G Bellemare},
  \bibinfo{person}{Alex Graves}, \bibinfo{person}{Martin Riedmiller},
  \bibinfo{person}{Andreas~K Fidjeland}, \bibinfo{person}{Georg Ostrovski},
  {et~al\mbox{.}}} \bibinfo{year}{2015}\natexlab{}.
\newblock \showarticletitle{Human-level control through deep reinforcement
  learning}.
\newblock \bibinfo{journal}{\emph{Nature}} \bibinfo{volume}{518},
  \bibinfo{number}{7540} (\bibinfo{year}{2015}), \bibinfo{pages}{529}.
\newblock


\bibitem[\protect\citeauthoryear{Moosavi-Dezfooli, Fawzi, Fawzi, and
  Frossard}{Moosavi-Dezfooli et~al\mbox{.}}{2017}]%
        {moosavi2016universal}
\bibfield{author}{\bibinfo{person}{Seyed-Mohsen Moosavi-Dezfooli},
  \bibinfo{person}{Alhussein Fawzi}, \bibinfo{person}{Omar Fawzi}, {and}
  \bibinfo{person}{Pascal Frossard}.} \bibinfo{year}{2017}\natexlab{}.
\newblock \showarticletitle{Universal adversarial perturbations}. In
  \bibinfo{booktitle}{\emph{CVPR}}.
\newblock


\bibitem[\protect\citeauthoryear{Moosavi-Dezfooli, Fawzi, and
  Frossard}{Moosavi-Dezfooli et~al\mbox{.}}{2016}]%
        {moosavi2016cvpr}
\bibfield{author}{\bibinfo{person}{Seyed-Mohsen Moosavi-Dezfooli},
  \bibinfo{person}{Alhussein Fawzi}, {and} \bibinfo{person}{Pascal Frossard}.}
  \bibinfo{year}{2016}\natexlab{}.
\newblock \showarticletitle{Deepfool: a simple and accurate method to fool deep
  neural networks}. In \bibinfo{booktitle}{\emph{CVPR}}.
\newblock


\bibitem[\protect\citeauthoryear{Nemati, Ghassemi, and Clifford}{Nemati
  et~al\mbox{.}}{2016}]%
        {nemati2016optimal}
\bibfield{author}{\bibinfo{person}{Shamim Nemati}, \bibinfo{person}{Mohammad~M
  Ghassemi}, {and} \bibinfo{person}{Gari~D Clifford}.}
  \bibinfo{year}{2016}\natexlab{}.
\newblock \showarticletitle{Optimal medication dosing from suboptimal clinical
  examples: A deep reinforcement learning approach}. In
  \bibinfo{booktitle}{\emph{Engineering in Medicine and Biology Society (EMBC),
  2016 IEEE 38th Annual International Conference of the}}. IEEE,
  \bibinfo{pages}{2978--2981}.
\newblock


\bibitem[\protect\citeauthoryear{Oh, Fritz, and Schiele}{Oh
  et~al\mbox{.}}{2017}]%
        {joon17iccv}
\bibfield{author}{\bibinfo{person}{Seong~Joon Oh}, \bibinfo{person}{Mario
  Fritz}, {and} \bibinfo{person}{Bernt Schiele}.}
  \bibinfo{year}{2017}\natexlab{}.
\newblock \showarticletitle{Adversarial Image Perturbation for Privacy
  Protection -- A Game Theory Perspective}. In
  \bibinfo{booktitle}{\emph{International Conference on Computer Vision
  (ICCV)}}.
\newblock


\bibitem[\protect\citeauthoryear{Pattanaik, Tang, Liu, Bommannan, and
  Chowdhary}{Pattanaik et~al\mbox{.}}{2017}]%
        {pattanaik2017robust}
\bibfield{author}{\bibinfo{person}{Anay Pattanaik}, \bibinfo{person}{Zhenyi
  Tang}, \bibinfo{person}{Shuijing Liu}, \bibinfo{person}{Gautham Bommannan},
  {and} \bibinfo{person}{Girish Chowdhary}.} \bibinfo{year}{2017}\natexlab{}.
\newblock \showarticletitle{Robust Deep Reinforcement Learning with Adversarial
  Attacks}.
\newblock \bibinfo{journal}{\emph{arXiv preprint arXiv:1712.03632}}
  (\bibinfo{year}{2017}).
\newblock


\bibitem[\protect\citeauthoryear{Peters, Vijayakumar, and Schaal}{Peters
  et~al\mbox{.}}{2003}]%
        {peters2003reinforcement}
\bibfield{author}{\bibinfo{person}{Jan Peters}, \bibinfo{person}{Sethu
  Vijayakumar}, {and} \bibinfo{person}{Stefan Schaal}.}
  \bibinfo{year}{2003}\natexlab{}.
\newblock \showarticletitle{Reinforcement learning for humanoid robotics}. In
  \bibinfo{booktitle}{\emph{Proceedings of the third IEEE-RAS international
  conference on humanoid robots}}. \bibinfo{pages}{1--20}.
\newblock


\bibitem[\protect\citeauthoryear{Pinto, Davidson, Sukthankar, and Gupta}{Pinto
  et~al\mbox{.}}{2017}]%
        {pinto2017robust}
\bibfield{author}{\bibinfo{person}{Lerrel Pinto}, \bibinfo{person}{James
  Davidson}, \bibinfo{person}{Rahul Sukthankar}, {and} \bibinfo{person}{Abhinav
  Gupta}.} \bibinfo{year}{2017}\natexlab{}.
\newblock \showarticletitle{Robust Adversarial Reinforcement Learning}. In
  \bibinfo{booktitle}{\emph{International Conference on Machine Learning}}.
  \bibinfo{pages}{2817--2826}.
\newblock


\bibitem[\protect\citeauthoryear{Sallab, Abdou, Perot, and Yogamani}{Sallab
  et~al\mbox{.}}{2017}]%
        {sallab2017deep}
\bibfield{author}{\bibinfo{person}{Ahmad~EL Sallab}, \bibinfo{person}{Mohammed
  Abdou}, \bibinfo{person}{Etienne Perot}, {and} \bibinfo{person}{Senthil
  Yogamani}.} \bibinfo{year}{2017}\natexlab{}.
\newblock \showarticletitle{Deep reinforcement learning framework for
  autonomous driving}.
\newblock \bibinfo{journal}{\emph{Electronic Imaging}} \bibinfo{volume}{2017},
  \bibinfo{number}{19} (\bibinfo{year}{2017}), \bibinfo{pages}{70--76}.
\newblock


\bibitem[\protect\citeauthoryear{Schaul, Quan, Antonoglou, and Silver}{Schaul
  et~al\mbox{.}}{2015}]%
        {schaul2015prioritized}
\bibfield{author}{\bibinfo{person}{Tom Schaul}, \bibinfo{person}{John Quan},
  \bibinfo{person}{Ioannis Antonoglou}, {and} \bibinfo{person}{David Silver}.}
  \bibinfo{year}{2015}\natexlab{}.
\newblock \showarticletitle{Prioritized experience replay}.
\newblock \bibinfo{journal}{\emph{arXiv preprint arXiv:1511.05952}}
  (\bibinfo{year}{2015}).
\newblock


\bibitem[\protect\citeauthoryear{Silver, Huang, Maddison, Guez, Sifre, Van
  Den~Driessche, Schrittwieser, Antonoglou, Panneershelvam, Lanctot,
  et~al\mbox{.}}{Silver et~al\mbox{.}}{2016}]%
        {silver2016mastering}
\bibfield{author}{\bibinfo{person}{David Silver}, \bibinfo{person}{Aja Huang},
  \bibinfo{person}{Chris~J Maddison}, \bibinfo{person}{Arthur Guez},
  \bibinfo{person}{Laurent Sifre}, \bibinfo{person}{George Van Den~Driessche},
  \bibinfo{person}{Julian Schrittwieser}, \bibinfo{person}{Ioannis Antonoglou},
  \bibinfo{person}{Veda Panneershelvam}, \bibinfo{person}{Marc Lanctot},
  {et~al\mbox{.}}} \bibinfo{year}{2016}\natexlab{}.
\newblock \showarticletitle{Mastering the game of Go with deep neural networks
  and tree search}.
\newblock \bibinfo{journal}{\emph{nature}} \bibinfo{volume}{529},
  \bibinfo{number}{7587} (\bibinfo{year}{2016}), \bibinfo{pages}{484--489}.
\newblock


\bibitem[\protect\citeauthoryear{Szegedy, Zaremba, Sutskever, Bruna, Erhan,
  Goodfellow, and Fergus}{Szegedy et~al\mbox{.}}{2014}]%
        {szegedy2014iclr}
\bibfield{author}{\bibinfo{person}{Christian Szegedy},
  \bibinfo{person}{Wojciech Zaremba}, \bibinfo{person}{Ilya Sutskever},
  \bibinfo{person}{Joan Bruna}, \bibinfo{person}{Dumitru Erhan},
  \bibinfo{person}{Ian Goodfellow}, {and} \bibinfo{person}{Rob Fergus}.}
  \bibinfo{year}{2014}\natexlab{}.
\newblock \showarticletitle{Intriguing properties of neural networks}. In
  \bibinfo{booktitle}{\emph{ICLR}}.
\newblock


\bibitem[\protect\citeauthoryear{Tesauro}{Tesauro}{1995}]%
        {tesauro1995td}
\bibfield{author}{\bibinfo{person}{Gerald Tesauro}.}
  \bibinfo{year}{1995}\natexlab{}.
\newblock \showarticletitle{Td-gammon: A self-teaching backgammon program}.
\newblock In \bibinfo{booktitle}{\emph{Applications of Neural Networks}}.
  \bibinfo{publisher}{Springer}, \bibinfo{pages}{267--285}.
\newblock


\bibitem[\protect\citeauthoryear{Van~Hasselt, Guez, and Silver}{Van~Hasselt
  et~al\mbox{.}}{2016}]%
        {van2016deep}
\bibfield{author}{\bibinfo{person}{Hado Van~Hasselt}, \bibinfo{person}{Arthur
  Guez}, {and} \bibinfo{person}{David Silver}.}
  \bibinfo{year}{2016}\natexlab{}.
\newblock \showarticletitle{Deep Reinforcement Learning with Double
  Q-Learning.}
\newblock


\bibitem[\protect\citeauthoryear{Wang, Schaul, Hessel, Hasselt, Lanctot, and
  Freitas}{Wang et~al\mbox{.}}{2016}]%
        {wang2016dueling}
\bibfield{author}{\bibinfo{person}{Ziyu Wang}, \bibinfo{person}{Tom Schaul},
  \bibinfo{person}{Matteo Hessel}, \bibinfo{person}{Hado Hasselt},
  \bibinfo{person}{Marc Lanctot}, {and} \bibinfo{person}{Nando Freitas}.}
  \bibinfo{year}{2016}\natexlab{}.
\newblock \showarticletitle{Dueling Network Architectures for Deep
  Reinforcement Learning}. In \bibinfo{booktitle}{\emph{International
  Conference on Machine Learning}}. \bibinfo{pages}{1995--2003}.
\newblock


\bibitem[\protect\citeauthoryear{Xie, Wang, Zhang, Zhou, Xie, and Yuille}{Xie
  et~al\mbox{.}}{2017}]%
        {xie2017adversarial}
\bibfield{author}{\bibinfo{person}{Cihang Xie}, \bibinfo{person}{Jianyu Wang},
  \bibinfo{person}{Zhishuai Zhang}, \bibinfo{person}{Yuyin Zhou},
  \bibinfo{person}{Lingxi Xie}, {and} \bibinfo{person}{Alan Yuille}.}
  \bibinfo{year}{2017}\natexlab{}.
\newblock \showarticletitle{Adversarial examples for semantic segmentation and
  object detection}. In \bibinfo{booktitle}{\emph{International Conference on
  Computer Vision. IEEE}}.
\newblock


\end{thebibliography}

\end{document}